\documentclass[10pt,twocolumn,letterpaper]{article}
\usepackage{multirow}
\usepackage{iccv}
\usepackage{times}
\usepackage{epsfig}
\usepackage{graphicx}
\usepackage{amsmath}
\usepackage{amssymb}
\usepackage{algorithm}
\usepackage{algpseudocode}
\usepackage{booktabs}
\usepackage{mathtools}
\usepackage{amssymb}

\usepackage[pagebackref=true,breaklinks=true,letterpaper=true,colorlinks,bookmarks=false]{hyperref}

\iccvfinalcopy %

\def\iccvPaperID{XX} %

\ificcvfinal\fi
\usepackage{cleveref}

\DeclarePairedDelimiter\norm{\lVert}{\rVert}

\begin{document}

\title{Distance Matters For Improving Performance Estimation Under Covariate Shift}

\author{Mélanie Roschewitz \\ 
Imperial College London\\
{\tt\small mb121@ic.ac.uk} 
\and Ben Glocker\\
Imperial College London\\
{\tt\small b.glocker@imperial.ac.uk} 
}

\maketitle
\ificcvfinal\thispagestyle{empty}\fi

\begin{abstract}
Performance estimation under covariate shift is a crucial component of safe AI model deployment, especially for sensitive use-cases. Recently, several solutions were proposed to tackle this problem, most leveraging model predictions or softmax confidence to derive accuracy estimates. However, under dataset shifts confidence scores may become ill-calibrated if samples are too far from the training distribution. In this work, we show that taking into account distances of test samples to their expected training distribution can significantly improve performance estimation under covariate shift. Precisely, we introduce a ``distance-check'' to flag samples that lie too far from the expected distribution, to avoid relying on their untrustworthy model outputs in the accuracy estimation step. We demonstrate the effectiveness of this method on 13 image classification tasks, across a wide-range of natural and synthetic distribution shifts and hundreds of models, with a median relative MAE improvement of 27\% over the best baseline across all tasks, and SOTA performance on 10 out of 13 tasks. Our code is publicly available at \url{https://github.com/melanibe/distance_matters_performance_estimation}. 

\end{abstract}

\section{Introduction}

\begin{figure}
    \centering
    \includegraphics{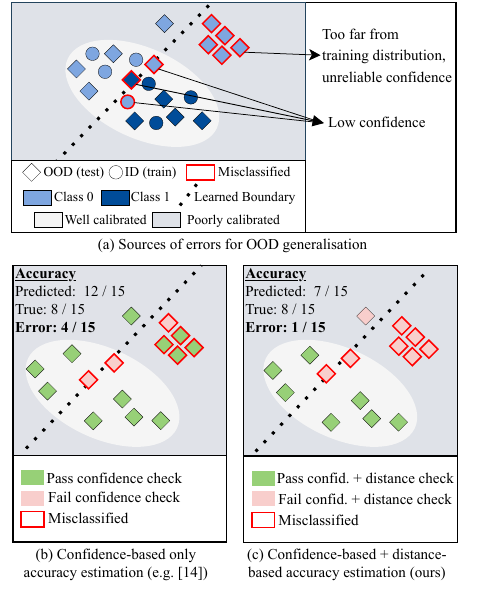}
    \caption{\textbf{Performance estimation under covariate shift needs to take into account different sources of errors.} Distance to the source distribution in the embedding space matters as confidence estimates become unreliable with increased distance.}
    \label{fig:fig1}
\end{figure}

Machine learning models are sensitive to variations in their deployment environments~\cite{Hendrycks_2021_ICCV,yu2021an,ovadia2019,koh2021wilds,wang2019learning,miller2021accuracy}. Due to the unavailability of ground truth labels for continuous performance monitoring at deployment time, real-time and accurate performance estimation is crucial to detect any unexpected behavior or model failure, particularly in distribution-shifted settings. This is especially important for sensitive use cases such as clinical decision making.
\\
    
The difficulty in estimating model performance arises from the lack of reliability of model outputs under covariate shift \cite{ovadia2019,ensemble}. Recently, several attempts have been made at addressing this problem, many of them based on confidence estimates~\cite{guillory2021predicting,garg2022leveraging,zeju}. For example, Average Thresholded Confidence (ATC)~\cite{garg2022leveraging} leverages softmax outputs for estimating classification accuracy, considering that all outputs whose confidence do not reach a certain threshold are incorrectly classified. While this method has shown to be effective at estimating the performance under mild shifts (e.g. on synthetically corrupted images), experiments show that the method under-performs in more substantial shifts such as natural sub-population shifts. In particular, current approaches tend to \emph{overestimate} accuracy in natural real-world distribution shift settings~\cite{garg2022leveraging,jiang2022assessing}. This can notably be explained by a deterioration of model calibration when going further from the training distribution \cite{kirsch2022a}, with softmax outputs becoming over-confident and unreliable~\cite{ovadia2019}. If test samples are too far from training samples, relying on the output of the classification layer for performance estimation is insufficient. From an uncertainty point of view, softmax outputs can been seen as capturing aleatoric uncertainty, arising from overlapping class boundaries~\cite{kendall2017uncertainties,chua2022tackling}. However, under dataset shifts, errors may also arise from the fact that the model has never seen this type of input data and does not know how to respond to such inputs. This is referred to as epistemic uncertainty \cite{chua2022tackling} and is not well captured by softmax outputs~\cite{kendall2017uncertainties,DUQ}, as demonstrated by its poor performance on the related out-of-distribution (OOD) detection task \cite{ovadia2019,ensemble,sun2022out,DUQ,lee2018simple}. Note that in OOD detection, the goal is to separate on separating ID from OOD inputs, regardless of the downstream classification performance, often considering inputs completely unrelated to the task. This differs from performance estimation under covariate shift, where we assume that the classification task still applies to the shifted inputs and we focus on estimating performance, not on detecting shifts.

\paragraph{Methodological contributions} In this paper, we argue that performance estimators should identify samples far away from the training set in the embedding space, for which softmax estimates are most likely unreliable. By measuring the distance in the embedding space, we are able to measure how well the model ``understood'' the sample when projecting the input to the classification space. This idea is illustrated in \cref{fig:fig1}. Following this intuition, we propose a simple yet effective method to improve the quality of current SOTA performance estimators. Specifically, we use nearest-neighbours distance in the embedding space to reject samples that lay too far from the training distribution. We then only use confidence-based performance estimators on the remaining samples, considering all previously rejected samples as mis-classified. Our distance check approach is versatile and can be used to improve the quality of various existing performance estimators (e.g. ~\cite{garg2022leveraging, jiang2022assessing}).

\paragraph{Main results} We evaluate our approach on 13 classification tasks ranging from cancer cell to animal classification. The nature of the distribution shifts studied covers a wide-range of shifts: from synthetic corruption, acquisition shift, real-world population shift to representation shift. For each task we evaluate between 18 and 259 models, covering various training strategies and network architectures. These experiments demonstrate that integrating distance into accuracy estimators significantly improves the quality of the estimation. For example, our proposed estimator ATC-DistCS is significantly better than previous SOTA ATC \cite{garg2022leveraging} on all but one task, with a median relative MAE improvement of 30\% across all tasks. Furthermore, comparing to the most recent COT method~\cite{lu2023predicting}, we demonstrate a 27\% median relative performance improvement across all tasks, with new SOTA performances on 10 out of 13 tasks. We also demonstrate significant improvements across all datasets for agreement based accuracy estimation when integrating our distance check. Ablation studies yield further insights in the method and its limitations. Finally, to the best of our knowledge, we provide the first comprehensive publicly available codebase of current SOTA baselines for accuracy estimation, along with the complete code to reproduce our experiments.

\section{Background}
\subsection{Performance estimation without ground truth}

Current methods for performance estimation under covariate shift can be broadly grouped in 4 categories:

\paragraph{Estimating performance via auxiliary task performance} Modifies the main classification model to incorporate a (sufficiently informative) auxiliary task for which ground truth labels are available at test time: accuracy on the main task is then approximated by the computed accuracy on the auxiliary task. For example, \cite{deng2021does} trains a multi-task model for predicting the class at hand as well as the rotation applied to the input image. The main limitation of this line of work is the requirement to build a multi-task model, making it unusable as a post-hoc tool.

\paragraph{Training a regressor between ID and OOD accuracy} This regressor can be trained based on model outputs or on measures of distance between datasets~\cite{elsahar2019annotate,schelter2020learning,guillory2021predicting,maggio2022performance,deng2021labels}. One major drawback of this class of estimators is their requirement for having access to labelled OOD data for training the regression model. This is not always available in practice, in particular in data-scarce domains such that healthcare. In absence of such OOD datasets, regressor are sometimes trained using corrupted versions of the original validation set as ``OOD" sets. However, this can not guarantee the robustness of this estimator against other shifts e.g. natural subpopulation shift~\cite{santurkar2021breeds}. 

\paragraph{Agreement-based estimators} Are based on the idea that agreement between member of model ensembles correlate with model accuracy. For example, generalised disagreement equality (GDE) \cite{jiang2022assessing} use pairs of models trained with different random seeds to compute disagreement. Others use more intricate methods for training specialised models to align disagreement and accuracy further~\cite{chuang2020estimating,chen2021detecting}. However, these procedures often require expensive additional training steps to derive the siblings models and are not applicable to post-hoc scenarios where only the final model is available to the end user. In \cite{baek2022agreementontheline}, the authors go as far as training dozens models to fit a regressor between agreement and accuracy, whereas \cite{chen2021detecting} requires training a new ensemble for every single test set requiring performance estimation. 

\paragraph{Confidence-based estimators} These methods, contrarily to the ones above, only require the final model's outputs to perform accuracy estimation and do not require any OOD data for calibration. As such, they are versatile and can be used with any classification model. For example, Difference of Confidence (DOC)~\cite{guillory2021predicting} approximates the difference in accuracy between the evaluation set and the in-distribution (ID) validation set by the difference in average model confidence. ATC~\cite{garg2022leveraging} introduces a confidence threshold such that all test samples for which the confidence is lower than this threshold are considered wrong and all samples meeting the minimum confidence requirement are considered correct (see Methods). Finally, concurrently to our work, COT~\cite{lu2023predicting} proposed to estimate accuracy based on based on optimal transport of model confidences. Precisely, they measure the Wasserstein distance between source label distribution and target softmax distribution to estimate the test error. Note, that this is expected to perform well if the source label distribution matches the target label distribution but might fail if this assumption breaks. 

\subsection{Distance-based out-of-distribution detection}
\label{sec:ooddetection}
The idea that OOD samples should lie far from the training samples in the embedding space is at the core of distance-based methods for OOD detection. For example, \cite{lee2018simple} propose to fit multi-variate Gaussians on the training embedding distribution of each class and use the Mahalanobis distance~\cite{mahalanobis1936generalised} to characterise how far test samples are from this expected distribution. If a sample is far from all class clusters, it is considered OOD. This method has shown some success at various OOD detection tasks~\cite{berger2021confidence, lee2018simple} and extensions of this work have since further improved its capabilities~\cite{ren2021simple}. However, this method suffers from one major limitation: it has a strong assumption that the class embeddings clusters can be accurately modelled by a Gaussian Multivariate distribution. Without any constraints on the training procedure or the embedding space at training time, this assumption may not hold~\cite{sun2022out}. This is the motivation for the work of~\cite{sun2022out} who proposed a non-parametric alternative OOD detection method. The authors still focus on the idea of using distances in the embedding space to detect OOD samples, but they leverage nearest-neighbours distances instead of the Mahalanobis distances, removing the normality assumption on the embedding. Precisely, they use the distance to the $\text{K}^{th}$ nearest-neighbour to classify samples as OOD. They derive the classification threshold for OOD versus ID task such that 95-99\% of the training samples are classified as ID.

\section{Methods}
In this section, we begin by reminding the reader of the core principles of two base performance estimators which we build on top of: ATC~\cite{garg2022leveraging} and GDE~\cite{jiang2022assessing}. We then introduce our plug-in distance checker designed to flag untrustworthy samples, and discuss how to incorporate this distance check into these performance estimators to yield our proposed estimators ``ATC-DistCS'' and ``GDE-DistCS''.

\subsection{Base estimators}
\paragraph{Average Thresholded Confidence (ATC)~\cite{garg2022leveraging}} approximates accuracy by the proportion of OOD predictions that do not exceed a certain confidence threshold (derived from the ID validation set, where confidence is defined as temperature-scaled~\cite{calibrationmodern} softmax confidence). Precisely, the threshold ATC is defined such that on source data $D_s$ the expected number of points that obtain a confidence less than ATC matches the error of the model on $D_s$. This method has been further refined by \cite{zeju}, where the authors propose to apply class-wise temperature scaling and to define class-wise confidence thresholds to improve the quality of the estimation, in particular for class-imbalanced problems.

\paragraph{Generalised Disagreement Equality (GDE)~\cite{jiang2022assessing}}  Assuming access to two models $g$ and $g'$ trained with different random seeds (but identical architecture and training paradigms), GDE estimates model accuracy by $\frac{1}{N}\sum_{i\in \text{test set}}\left[g(x_i) = g'(x_i)\right]$, where N is the size of the OOD test set, $x_i$ the inputs to the model, and $g(x_i)$ denotes the model prediction. 

\subsection{Integrating distance to training set}
\paragraph{Average Distance Check} Inspired by the OOD detection work in \cite{sun2022out}, we propose to improve standard performance estimators with a ``distance checker". Instead of simply rejecting samples with low confidence or model disagreement, we argue that the distance of any given sample to the in-distribution training set should also be taken into account to determine whether its confidence (and prediction) is likely to be trustworthy for estimation purposes or not. In simple terms, we ``reject'' samples whose penultimate-layer embeddings lie in a region ``far'' from the ID embedding space. The distance from a sample to the in-distribution set is determined by the average distance between the sample and all of its K-nearest-neighbours: 
\begin{equation}
    \text{AD}_i = \frac{1}{K}\sum_k \norm*{f_i - n_i^{(k)}}_2,
\end{equation} where $K$ is the number of nearest-neighbours to consider, $f_i$ is the embedding of the $i^{th}$ test sample and $n_i^{(k)}$ the $k^{th}$ nearest neighbours to $f_i$ in the embedding space, nearest neighbours are searched for in the training set. The acceptable threshold is determined on the in-distribution validation set as the $99^{th}$-percentile of the average distances observed on this set i.e.
\begin{equation}
    \text{DistThreshold} = \text{quantile}_{.99}\left \{AD_i, \forall i \in \text{val set}\right\}\text{.}
\end{equation}
 Note, that our distance criterion differs from \cite{sun2022out}, in that (i) we use the average of all K distances instead of the distance to the $\text{K}^{th}$ neighbour only (to be less sensitive to outliers); (ii) we do not normalise the embeddings (see ablation study in \cref{sec:results}); (iii) we do not use a contrastive loss for training our models as this assumption may restrict the scope of application of the method. The fitting procedure for the distance checker can be found in \cref{alg:our}.

\paragraph{Using distance to improve the quality of performance estimators} Our proposed ``distance-checker" can be used as a plug-in method to improve the estimation results of different existing accuracy estimators. Specifically, first we propose ``ATC-Dist'', where we combine both criteria to estimate the accuracy under shift: a sample is estimated as being correct if it is (i) of high enough confidence, (ii) not too far from the in-distribution embeddings. Similarly we extend GDE with our distance criterion to get ``GDE-Dist''. There the correctness of a sample is estimated by (i) agreement between both models, and (ii) distance to the in-distribution embeddings. The estimation procedure for ATC-Dist and GDE-Dist is shown in \cref{alg:our}.

\begin{algorithm}
\caption{}\label{alg:our}
\begin{algorithmic}
\Procedure{Fit DistanceChecker}{$X_{train}$, $X_{val}$}
\State $f_{train} \gets \Call{get features}{X_{train}}$
\State $f_{val} \gets \Call{get features}{X_{val}}$
\State KNN $\gets \Call{Fit Nearest Neighbors}{f_{train}}$
\State $\text{AD}_{val} \gets \Call{Average NN Distances}{\mbox{KNN}, f_{val}} $
\State DistThreshold $\gets \Call{Quantile}{\text{AD}_{val}, .99}$
\State $\textbf{return}$ DistThreshold, KNN 
\EndProcedure
\\
\Procedure{Get ATC-DIST}{$X_{\text{test}}$, ATC, KNN, DistThreshold}
\State $f_{\text{test}} \gets \Call{get features}{X_{\text{test}}}$
\State $\text{AD}_{\text{test}} \gets \Call{Average NN Distances}{\mbox{KNN}, f_{\text{test}}}$ 
\State $\text{kept}_{\text{Dist}} \gets \text{AD}_{\text{test}} < \text{DistThreshold}$ \Comment{Distance check}
\State $\text{c}_{\text{test}} \gets \Call{softmax confidence}{X_{\text{test}}}$
\State $\text{kept}_{\text{ATC}} \gets \text{c}_{\text{test}} > \text{ATC} $ \Comment{Confidence check}
\State ATC-DIST $ \gets \frac{|\text{kept}_{ATC} \ \cap \ \text{kept}_{\text{Dist}}|}{|X_{\text{\text{test}}}|}$
\State $\textbf{return}$ ATC-DIST 
\EndProcedure
\\
\Procedure{Get GDE-DIST}{$X_{\text{test}}$, $g_1$, $g_2$, DistThreshold, KNN}
\Comment{$g_1$, $g_2$ two models trained with different seeds}
\State $f_{\text{test}} \gets \Call{get features}{g_1, X_{\text{test}}}$
\State $\text{AD}_{\text{test}} \gets \Call{Average NN Distances}{\mbox{KNN}, f_{\text{test}}}$ 
\State $\text{kept}_{\text{Dist}} \gets \text{AD}_{\text{test}} < \text{DistThreshold}$ \Comment{Distance check}
\State $\text{agree}_{\text{\text{test}}} \gets g_1(X_{\text{\text{test}}}) = g_2(X_{\text{\text{test}}})$ \Comment{Agreement}
\State GDE-DIST $ \gets \frac{|\text{agree}_{\text{test}} \ \cap \ \text{kept}_{\text{Dist}}|}{|X_{\text{test}}|}$
\State $\textbf{return}$  GDE-DIST 
\EndProcedure
\end{algorithmic}
\end{algorithm}

\paragraph{Class-wise distance thresholds} As the tightness of class clusters may differ for different classes, we argue that the quality of the distance threshold can be further improved by defining class-wise distance thresholds. Concretely, for each class $c$ we compute $\text{DistThreshold}_c$ by taking the $\text{99}^{th}$ percentile of the average distance distribution of the subset of cases labelled as $c$ in the validation set. At test time, we use the distance threshold associated with the predicted class to determine the validity of a given sample prediction. In cases where less than 20 samples were present in the validation set for any given class, we use the global threshold for this class. Replacing the global distance threshold by the classwise thresholds in the procedure described above, yields our proposed ``ATC-DistCS'' and ``GDE-DistCS'' estimators.

\section{Results}
\label{sec:results}

\begin{figure*}
    \centering
    \includegraphics[width=0.9\linewidth]{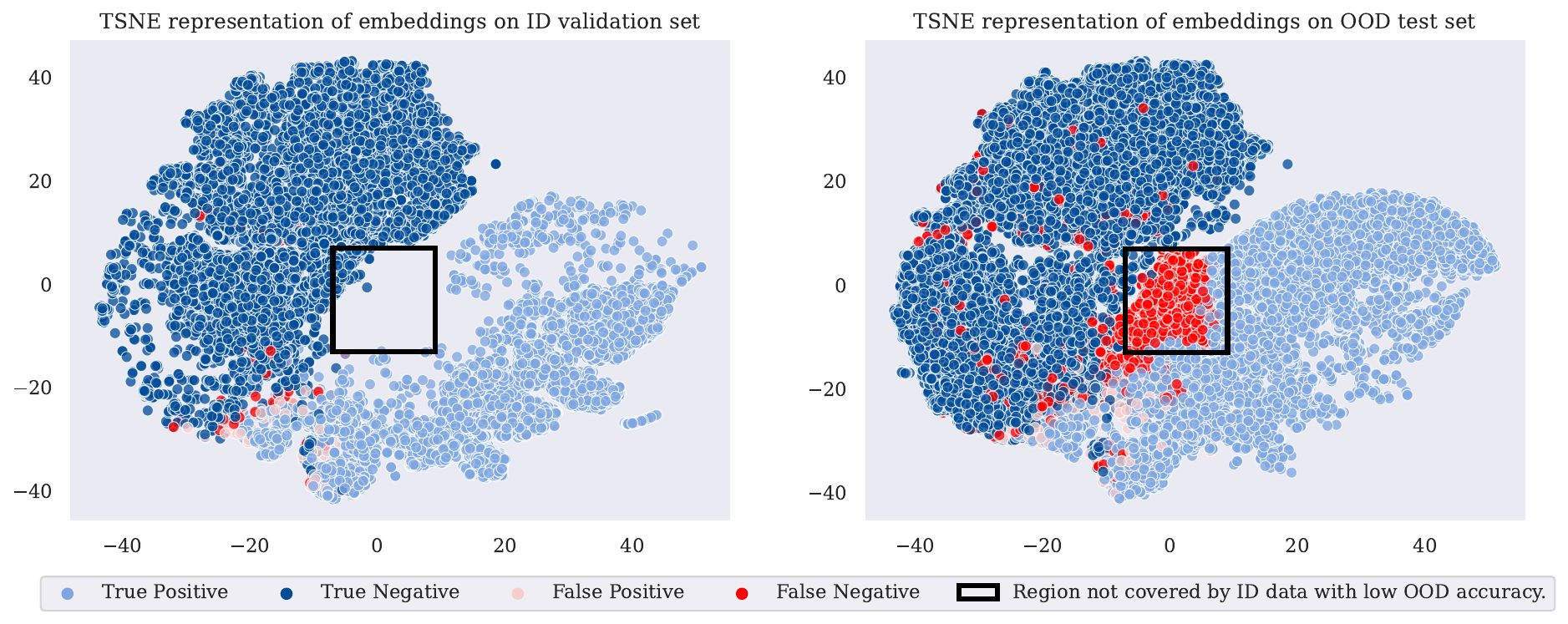}
    \caption{\textbf{Why distance matters, an example.} Joint TSNE~\cite{tsne} representation of the ID validation set and OOD test set plotted separately for a ResNet18 model on the WILDS CameLyon dataset. We can clearly distinguish a region with low density on the validation set and high density on the OOD set, where most points are misclassified. }
    \label{fig:tsne}
\end{figure*}

\paragraph{Motivating example} Prior to diving into quantitative analysis of our results, let's start with an illustrative example of our main idea: \textit{in the embedding space, regions of the shifted test set not covered by the ID set are likely to be regions of very low accuracy}. This pattern appears distinctively in the example in \cref{fig:tsne}, where we show the T-SNE~\cite{tsne} representation of the embeddings of the ID validation set as well as the OOD test set, on a model trained on the WILDS-CameLyon~\cite{koh2021wilds, bandi2018detection} dataset. We can clearly see how the region in black $-$ which is not well represented in the ID validation set $-$ contains an extremely high proportion of errors in the OOD test set.

\paragraph{Datasets} We validate our proposed method on a wide range of tasks and covering various natural and synthetic distribution shifts (more details in supplement): 
\begin{itemize}
    \item ImageNet~\cite{ILSVRC15} to ImageNet-Sketch~\cite{wang2019learning} where the distribution shifts from photographs to sketch; ImageNet-A~\cite{hendrycks2021natural}, where the distribution shifts adversarially; ImageNet-V2~\cite{recht2019imagenet} a setting with only mild shifts, designed to mimic ImageNet test set.
    \item CIFAR10~\cite{krizhevsky2009learning} to CIFAR10-C~\cite{hendrycks2018benchmarking} covering various synthetic corruptions, yielding 95 OOD datasets. 
    \item MNIST~\cite{lecun1998gradient} to SVHN~\cite{netzer2011reading}, classic digit classification, shifting from binary digit images to house numbers. 
    \item WILDS~\cite{koh2021wilds} benchmark, designed to study natural shifts occuring ``in the wild''. WILDS-Camelyon17~\cite{bandi2018detection} defines a histopathology binary task, with staining protocol shifts. WILDS-iCam~\cite{beery2020iwildcam} is a 182-classes animal classification task from camera traps, with shifts in camera location. WILDS-FMoW~\cite{christie2018functional} is a satellite image 62-class task, with temporal and geographical shifts. WILDS-RxRx1~\cite{taylor2019rxrx1} is a 1,139 genetic treatment classification task on fluorescent microscopy images, where the shift occurs from so-called experimental ``batch-effect''.
    \item The BREEDS~\cite{santurkar2021breeds} benchmark defines various tasks (Entity30, Entity13, Living17, NonLiving26) based on ImageNet subsets and superclasses. The main task consisting of predicting the super-class and the train-val-test split defined such that the subpopulations covered by the OOD test set are disjoint from the ones represented in the training and validation set. 
    \item PACS~\cite{li2017deeper} a 7-class task, where models are trained and validated on photographs and tested on 3 other domains (painting, sketches and cartoon). 
    \item PathMNIST~\cite{yang2023medmnist} histopathology 9-class task, where training and test splits are taken from different sites. 
\end{itemize}

\paragraph{Experimental setup and models} For each evaluated model, we fit our nearest neighbours algorithm on the training set, using K=25 neighbours for the distance check. Distance thresholds are computed on the in-distribution validation set. For each task, we evaluate our accuracy estimator on all available OOD sets, as described above, and measure estimation quality in terms of Mean Absolute Error (MAE) between predicted and true accuracy across all models. Note that if there were more than N=50,000 training samples, we randomly subsampled N samples in the K-NN fitting step to speed up inference. Moreover, for ImageNet to avoid doing a full inference pass on the extremely large training set, we fitted the K-NN algorithm directly on the validation set (discarding distance to self to get the distance threshold). For each task, we evaluate the quality of performance estimation on a large variety of models. For ImageNet, we test on 259 pretrained models from the \texttt{timm}~\cite{rw2019timm} package, covering a range of 14 family of model architectures. For all other datasets, we trained models ourselves using various architectures, training setups, random seeds and initialising models both from ImageNet and random weights (except for BREEDS datasets as they are build from ImageNet images, hence pretrained weights would violate the OOD assumptions of the testing subpopulations); amounting to 18 models for BREEDS tasks and 30 models for all other tasks. More details can be found in Supp Note 2 and in our codebase.

\paragraph{Choice of baselines} Our first analysis focuses on single-model accuracy estimation. We compare our method to established ATC~\cite{garg2022leveraging} and DoC~\cite{guillory2021predicting} baselines as well as their improved class-wise version \cite{zeju}. For class-wise estimation, if any given target class was not present in the validation set, or if less than 20 samples were predicted for that class, we used the global temperature and ATC-threshold for that particular class. This may happen for some classes in imbalanced datasets or with an extremely high number of classes (e.g. WILDS RxRx1 or WILS iCam). We also compare our method to the recently proposed COT estimator~\cite{lu2023predicting}. Note that, to date, in their pre-print, the authors only tested their method on a very limited set of tasks, as such our evaluation considerably extends the assessment of COT's capabilities. Regression-like methods are not included as we assume that no OOD dataset is available at training and validation time, similarly we do not include methods that require external metadata such as Mandoline~\cite{chen2021mandoline} as it was not available. Methods such as self-training ensembles~\cite{chen2021detecting} which require model retraining for every single test set, were also not considered as they were computationally much more heavy (it would require training over 3,000 ensembles in our experimental setting) and do not allow for real time monitoring. Weaker baselines such that simply using the average softmax confidence as accuracy estimation are not included as the extensive analysis in the ATC paper \cite{garg2022leveraging} already clearly demonstrates the superiority of ATC as a baseline. In a second analysis, we place ourselves in the scenario where we have access to two models for each task for accuracy estimation and compare agreement-based estimator GDE~\cite{jiang2022assessing} to our improved version GDE-DistCS. %

\begin{table*}\centering
\begin{tabular}{@{}lccccccccc@{}}\toprule
 & \multicolumn{4}{c}{Global TS \& ATC thresholds} & \phantom{ab}& \multicolumn{4}{c}{Classwise TS \& ATC thresholds}\\
\cmidrule{2-6} \cmidrule{8-10}
&  DoC & COT & ATC & ATC-Dist & ATC-DistCS & & COT & ATC & ATC-DistCS \\ 
Dataset family &  \cite{guillory2021predicting} & \cite{lu2023predicting} & \cite{garg2022leveraging} & (ours) & (ours) &  & ~\cite{lu2023predicting} & ~\cite{zeju} & (ours) \\ 
\midrule
ImageNet-Sketch & 19.26** & 4.62** & 6.05** & 4.75** & \textbf{3.28} & & 5.31** & 5.36** & \textbf{3.42}\\
ImageNet-A & 38.66** & 31.73** & 26.81** & 23.20** & 17.95** & & 21.77** & 35.67** & \textbf{15.10}\\
ImageNet-V2 & 4.94** & 5.70** & 1.90** & 1.43** & \textbf{0.64} & & 1.83** & 5.39** & 3.51** \\
Living17 & 21.08** & 18.95** & 17.98** & 15.86** & 14.45* & & 20.18** & 15.02** & \textbf{11.82}\\
NonLiving26 & 24.31** & 21.38* & 16.71** & 15.65** & \textbf{14.53} &  & 21.85* & 15.84** & \textbf{13.87} \\
Entity13 & 13.55** & 12.78** & 8.96** & 8.30* & \textbf{8.15} & & 12.99** & 8.64** & \textbf{7.84} \\
Entity30 & 17.75** & 15.45** & 12.31** & 11.65* & \textbf{11.31} & & 15.98** & 12.15** & \textbf{11.15}\\
WILDS CameLyon & 7.57** & 3.07** & 6.86** & 4.90** & 4.71** & & \textbf{2.99} & 6.82** & 4.69** \\
WILDS iCam & \textbf{8.14} & 7.72* & \textbf{7.15} & 7.92* & 9.13** &  & \textbf{6.48} &  \textbf{5.39} &  \textbf{6.95} \\
WILDS FMoW & 3.54** &  2.04** & 2.72** & 2.06* & \textbf{1.91} & & 1.94* & \textbf{1.36} & \textbf{1.58} \\
WILDS RxRx1 & 7.47** & \textbf{2.36} & 6.02** & 5.01** & 3.86** & & \textbf{2.54} & 8.87** & 9.62** \\
MNIST & 61.41** & \textbf{15.17} & 49.52** & \textbf{17.41} & \textbf{15.96} & & \textbf{15.33} & 41.44** & \textbf{16.12} \\
PACS & 55.38** & \textbf{12.61} & 45.98** & 26.25** & 26.21**  & & 13.23* & 49.45** & 26.65** \\
PathMNIST & 3.68** & 9.90** & 2.67** & \textbf{1.31} & \textbf{1.14} & & 9.92** & 2.37** & \textbf{1.09} \\
CIFAR10 & 2.73** &  1.53** & 1.20** & 1.11*  & \textbf{1.08} & & 1.59** & 1.24** & \textbf{1.07} \\
\bottomrule
\end{tabular}
\caption{\textbf{Improving confidence-based accuracy estimation - summary table.} Results are reported in terms of Mean Absolute Error (in \%) across all models and OOD datasets. * denotes a p-value after Bonferroni correction \textless 0.05, ** p-value \textless 1e-3 for Wilcoxon signed-rank test~\cite{Wilcoxon} to test for difference between the best method versus all the others, for each dataset.  Bold denotes the best model, all methods not significantly different from the best are highlighted.}
\label{tab:main_res}
\end{table*}

\begin{figure}
    \centering
    \includegraphics[width=0.7\linewidth]{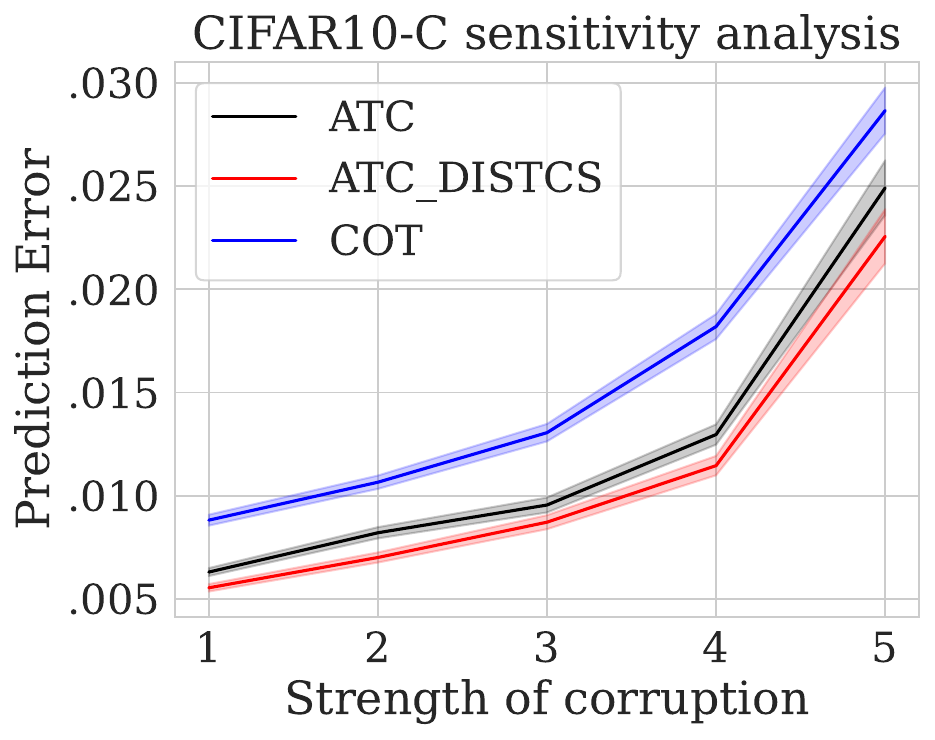}
    \caption{\textbf{Ablation study MSE in function of corruption strength for CIFAR10-C} across all models, shaded area depicts +/- one standard deviation.}
    \label{fig:cifar}
\end{figure}

\begin{table}\centering
\begin{tabular}{@{}rrrr@{}}\toprule
 & GDE~\cite{jiang2022assessing} & +DistCS \\
 Dataset family & & (ours) \\
\midrule
Living17 & 19.92** & \textbf{16.60} \\
NonLiving26 & 23.49** & \textbf{21.26} \\
Entity13 &  12.62** & \textbf{11.78} \\
Entity30 & 16.57** & \textbf{15.67} \\
WILDS CameLyon & 4.96** & \textbf{3.62} \\
WILDS iCam & \textbf{6.44} & \textbf{5.93} \\
WILDS FMoW &  9.39** & \textbf{8.56} \\
WILDS RxRx1 & 9.14** & \textbf{7.78} \\
MNIST & 34.63**  & \textbf{15.63} \\
PACS & 42.41** & \textbf{27.76} \\ 
PathMNIST & 2.81** & \textbf{1.62}  \\
CIFAR10 & 4.73** & \textbf{4.19} \\
\bottomrule
\end{tabular}
\caption{\textbf{Results for improving agreement-based estimates.} Best in bold, * denotes a p-value \textless 0.05, ** \textless 1e-5 for Wilcoxon-signed-rank test~\cite{Wilcoxon} for GDE-DistCS against GDE.}
\label{tab:gde_results}
\end{table}

\paragraph{Results for single-model performance estimation}
In \cref{tab:main_res}, we compare DoC, ATC, COT and our method in two different settings, one where temperature scaling (TS)~\cite{calibrationmodern} and ATC threshold are optimised globally for the entire dataset (left column group) and the second where we apply class-wise TS and ATC thresholds (rightmost columns). Temperature scaling is applied as previous studies have shown better results over raw model ouputs~\cite{zeju,garg2022leveraging}. Results are presented in terms of MAE over all shifted test sets and all models for each task. We can see that our method ATC-Dist achieves lower MAE than its counterpart ATC across all but one dataset (Wilds iCam, see discussion section). Furthermore, on all these datasets, ATC-DistCS using class-wise distance thresholds  further improves the results over ATC-Dist. The overall median relative MAE reduction of ATC-DistCS over all datasets is of 30\% compared to standard ATC in the global setting and 13\% in the class-wise setting. Additionally, our experimental results confirm the preliminary findings of~\cite{zeju} i.e. ATC with class-wise optimisation of temperature and thresholds outperforms ATC with global optimisation for the majority of datasets (only performance on heavily imbalanced CIFAR10-C had been reported so far). To measure statistical significance, for each dataset, we use the Wilcoxon signed-rank test~\cite{Wilcoxon} to compare the best method (i.e. with lowest MAE on that task) against all other methods, with Bonferroni~\cite{bonferroni1936teoria} correction to account for multiple testing. We highlight in bold the method with lowest MAE and all methods that are not significantly different to this method (at the level 0.05 after correction). We can see that ATC-DistCS achieves SOTA results on 10 out of 13 tasks, with a median relative MAE improvement of 27\% for ATC-DistCS over COT with global TS and 30\% with class-wise TS. We discuss differences between COT and ATC-DistCS in more details in \cref{sec:discussion}. Finally, we detail the performance comparison on CIFAR10-C in \cref{fig:cifar}, we can see that our method outperforms the baselines at all levels of corruption strength. Additional scatterplots detailing predicted versus true accuracy can be found in Supp. Note 3. %

\paragraph{Results for agreement-based accuracy estimation.}
Our distance check is not only tailored for improving ATC but rather is a  general addition that can be ``plugged-in'' to various estimators. We demonstrate this by showing how our method improves the quality of agreement-based estimator GDE~\cite{jiang2022assessing}, another well established baseline. For every training configuration, we repeat training with 3 different seeds. To estimate the accuracy for model $g_1$, we use another model $g_2$ trained with a different seed to compute disagreement and deduce the predicted accuracy for $g_1$. We then further improve the estimation with our proposed distance check i.e. we fit our distance checker to the validation features on $g_1$ and use it on the corresponding OOD features to discard distant samples. We evaluate the error for every model using all possible pairs. Results are summarised in \cref{tab:gde_results}. The proposed GDE-DistCS shows statistically significant improvements across all tasks (expect for one task where it is equivalent), with a median relative MAE improvement of 13\% over the standard GDE method. However, it is worth noting that results obtained  with the accuracy estimators from the previous paragraph are systematically better than these disagreement estimates.

\begin{figure*}
    \centering
    \includegraphics[width=.87\linewidth]{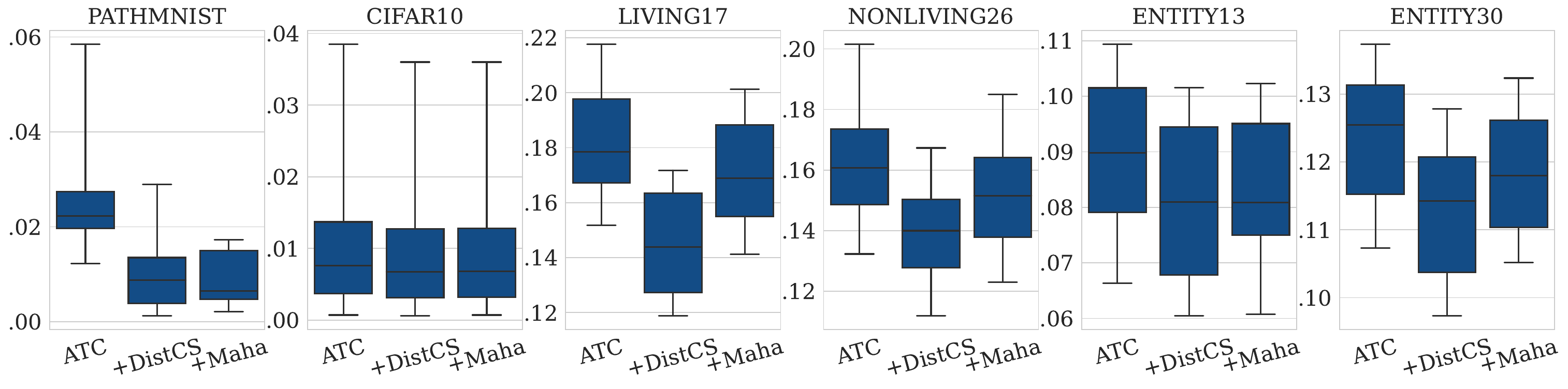}
    \caption{\textbf{Ablation study for the choice of distance estimation method: K-NN (DistCS) versus Mahalanobis distance (Maha).} Each boxplot shows the distribution of the Mean Absolute Error for accuracy estimation. Whiskers denote the [5\%;95\%]-percentiles of the distribution, outliers omitted for readability. Using distance improves the results for all but one dataset, no matter if K-NN or Mahalanobis distance. However, K-NN distance is better than Mahalanobis overall. For additional datasets, see Supp. Note 4.}
    \label{fig:ablation_distance}
\end{figure*}

\paragraph{Ablation studies: choice of distance measure and K-NN hyperparameters} We ran additional experiments to justify our choice to use K-NN distance for detecting unreliable samples. As mentioned in \cref{sec:ooddetection}, other methods have been proposed to perform distance-based OOD detection. Most famous is the Mahalanobis criteria proposed by~\cite{lee2018simple}. Hence, we compare the performance of the proposed ATC-DistCS to ATC-Maha where we use Mahalanobis to compute the distance (all other steps the same). Results in \cref{fig:ablation_distance} show that (i) for most  datasets adding the distance check helps, regardless of the distance choice; (ii) the K-NN distance performs better than Mahalanobis distance (and is often computationally faster). Secondly, in Supp Note 5, we also investigate the impact of the number of neighbours and the effect of features normalisation (as it improved OOD detection in~\cite{sun2022out}), showing that our method is robust to the choice of number of neighbours and does not require normalisation. Similarly, Supp. Note 6 shows that our $99^{th}$-percentile distance threshold choice for rejecting samples is generalisable across all datasets, alleviating the need for cumbersome hyperparameter tuning and allowing us to all parameters fixed across all experiments.

\section{Discussion \& Conclusion}
\label{sec:discussion}
\paragraph{Main take-aways} The proposed ``distance-check'' significantly improves accuracy estimation results across datasets and tasks both for ATC and GDE;  with a median relative MAE improvement of 30\% for ATC versus ATC-DistCS in the global setting (resp. 13\% in the class-wise setting) and 13\% for GDE versus GDE-DistCS. Importantly, our method is versatile, can be applied to any model and does not require any OOD data to tune the performance estimator. In particular, we can apply the method even if we only have access to the final model at deployment time, enabling external performance monitoring (e.g. by regulators or local auditing teams). This contrasts with some recent methods that require dozens of models to improve upon ATC results~\cite{baek2022agreementontheline}. Moreover, we would like to underline the demonstrated plug-in aspect of the proposed method, i.e. its ability to improve estimation quality across several ``base" accuracy estimators. Indeed, this attests that distance to the expected distribution has to be taken into account for improved performance estimation and that estimators should not solely rely on model outputs. This is further corroborated by our ablation study comparing the use of K-NN versus Mahalanobis distances for the distance check step in the proposed accuracy estimation flow. Indeed, results show that regardless of the distance measure choice, our proposed ATC with distance check outperforms the standard ATC baseline for all but one dataset. This work is, to the best of our knowledge, the first proposing to combine confidence and distance based performance estimation, without requiring access to OOD data at calibration time. 

\paragraph{ATC-DistCS, COT and computational considerations} Results show that our method performs significantly better (or equivalent) to the concurrently proposed COT method on 10 out of 13 datasets. Our extensive evaluation not only justifies our method but also allows to gain more insight into COT, as it had only be evaluated on a few tasks in the original work~\cite{lu2023predicting}. Another important consideration is that COT's runtime increases in $\mathcal{O}(n^3)$ with the number of test samples and linearly with the number of classes, whereas K-NN distance increases linearly with the number of training samples (here limited to 50,000). The authors in \cite{lu2023predicting} propose to alleviate this problem by splitting the test set in batches and averaging accuracy estimates. Despite following this and limiting the number of test samples to 25,000, we still observed a runtime penalty of approximately one order of magnitude compared to ATC-DistCS for datasets with a high number of classes and where the transport optimisation problem needed a large number of iterations before convergence (e.g. on our CPU for a ImageNet ResNet150 model it took over 3500s to get one COT estimate versus 300s for ATC-Dist, 450s vs 50s for Wilds-RR1). Finally, the lightweight aspect of ATC-DistCS (and the ATC baseline) is in start contrast with other proposed methods such as e.g. self-training ensembles~\cite{chen2021detecting} which requires training new ensembles for every single evaluation set, highly impractical in real-world monitoring scenarios.

\paragraph{Limitations} The proposed method relies on the representativeness of the in-distribution validation set to calibrate the distance threshold. In other words, the validation set should cover the expected set of possibilities encountered in-distribution. If the validation data is not sufficiently representative of the ID setting e.g. not all classes are represented in the validation set, then the distance check is expected to be sub-optimal. This is what is happening with the WILDS-iCam results in the section above. For this heavily data imbalanced task, not all possible targets are present in the validation set. This led to the distance check not improving the results due to a sub-optimal distance threshold choice. Moreover, because classes were missing in the validation set we were not able to compute class-wise thresholds for many classes and had to use the global threshold for these classes, which are especially important in heavily imbalanced settings, as argued by \cite{zeju}. Similarly, in Wilds-RxRx1 many classes had only a few samples in the given in-distribution validation set leading to sub-optimal classwise thresholds for this equally imbalanced task. 

Finally, our method, by design, generates more conservative performance estimates than their counterparts without the distance check. As it considers that any point that lies too far from the expected embedding space is wrong, it will reduce the estimated accuracy. In most cases, this assumption holds in practice as shown by our experimental results. However, in some settings with extremely heavy distributional shift such as PACS, this assumption may lead to rejecting a lot of samples that appear ``too'' OOD. This may in turn yield excessively conservative performance estimates. Nevertheless, we argue that in practice if the input data is very far from the expected training distribution, having a low accuracy estimate triggering an auditing alert is a more desirable behaviour than having over-confident accuracy estimates which may mislead the user and generate unsafe AI use. Once the system is validated on the new data, it can easily be included in the calibration set. 

\paragraph{Conclusion}
Taking into account distance to the training distribution substantially improves performance estimation on a wide-range of tasks. Our proposed estimators implementing a distance check demonstrate SOTA performance on a large variety of tasks and significant improvement over previous SOTA baselines. Our method offers a practical and versatile approach to performance estimation on new data distributions, and thus, enables important safety checks for AI model deployment in critical applications. Importantly, our work clearly demonstrates the need to bridge the gap between performance estimation and traditional OOD detection literature and proposes a first step towards this end.

\section{Acknowledgements}
M.R. is funded through an Imperial College London
President’s PhD Scholarship. B.G. received support from the Royal Academy of
Engineering as part of his Kheiron/RAEng Research Chair.

{\small
\bibliographystyle{ieee_fullname}
\bibliography{bib}
}

\newpage
\onecolumn
{\Large\textbf{\centering Supplementary material for: \\Distance Matters For Improving Performance Estimation Under Covariate Shift}}

\section*{Supp Note 1: Additional information on datasets used for this study}
\begin{table}[ht]\centering
\begin{tabular}{@{}ccccc@{}}\toprule
Dataset family & ID datasets / splits & OOD datasets / splits & N classes & Type of shift \\ 
\midrule
ImageNet & Train / Val & ImageNet-Sketch & 1,000 & Photographs to Sketches \\
\multirow{3}{*}{PACS} & \multirow{3}{*}{Photo train / val} & Painting test & \multirow{3}{*}{7} & \multirow{3}{*}{Art type} \\
 & & Cartoon test &  &  \\
 & & Sketches test &  &  \\
MNIST & MNIST & SVHN & 10 & Binary images to house numbers \\
CIFAR10 & Train / Val & CIFAR10-C & 10 & Synthetic corruptions \\
Entity30 & Train / Val & OOD test & 30 & Subpopulation shift (random) \\
Entity13 & Train / Val & OOD test & 30 & Subpopulation shift (random) \\
Living17 & Train / Val &  OOD test & 17 & Subpopulation shift (random) \\
NonLiving26 & Train / Val & OOD test & 26 & Subpopulation shift (random) \\
WILDS Camelyon & Train / id-Val & ood-test, ood-val & 2 & Site / staining protocol \\
WILDS iCam &  Train / id-Val & ood-test, ood-val & 182 & Location of camera \\
WILDS FmoW & Train / id-Val & ood-test, ood-val & 62 & Location and time \\
WILDS RxRx1 & Train / id-Val & ood-test, ood-val & 1,189 & Experimental session \\
PathMNIST & Train / Val & Test & 9 & Site / staining protocol \\
\bottomrule
\end{tabular}
\caption{\textbf{Additional information for the dataset used.} ``Dataset family'' denotes the name used in the tables in the main paper to refer to this task. For all datasets we used the official splits as denoted in the columns.}
\label{tab:add}
\end{table}

\section*{Supp Note 2: Additional information on model training}
All our training and evaluation code as well as data augmentation and training configurations are available in our codebase \url{https://github.com/melanibe/distance_matters_performance_estimation}.

\paragraph{ImageNet models} For ImageNet, we used readily available trained models from the \texttt{timm}~\cite{rw2019timm} package. We evaluate all available models from the following 14 family of model architectures: ConvNext~\cite{liu2022convnet}, ConvMixer~\cite{trockman2022patches}, DarkNet~\cite{redmon2017yolo9000}, CSPNet~\cite{wang2020cspnet}, EfficientNet~\cite{tan2019efficientnet}, Inception ResNet~\cite{szegedy2017inception}, ResNext~\cite{xie2017aggregated}, ResNeSt~\cite{zhang2022resnest}, TResNet~\cite{ridnik2021tresnet}, DenseNet~\cite{huang2017densely}, ResNet~\cite{he2016deep}, ResNetv2~\cite{szegedy2017inception}, ECA-Net~\cite{wang2020eca}, Res-SE-Net~\cite{varshaneya2021res}. This amounted to testing a total of 259 trained models. 

\paragraph{Trained models}
For each model / training configuration we repeated training for 3 different seeds. Details of training configurations are listed in the table below. In total, we train 18 models from random initialisation and 12 models from ImageNet weights, amounting to 30 models for each dataset, except for the BREEDS datasets for which we only used the models trained from scratch (as they are subsets of ImageNet). The ``standard'' training procedure uses Adam optimiser, automatic learning rate adaptation after 10 epochs without improvement, early stopping when the accuracy did not improve anymore for 15 epochs. Unless specified otherwise, we used data augmentation during training (incl. random rotation, color jittering, flips, cropping), all data augmentations configurations can be found in the codebase. 

\begin{table}[]
    \centering
    \begin{tabular}{@{}llll@{}}
    \toprule
        Model architecture & Type of init. & Training procedure & Data augmentation \\
    \midrule
        ResNet18~\cite{he2016deep} & Random & Standard & Yes \\
        ResNet18~\cite{he2016deep} & Random & Standard + Weight decay $1e-3$ & Yes \\
        ResNet18~\cite{he2016deep} & ImageNet weights & Standard & Yes \\
        ResNet50~\cite{he2016deep} & Random & Standard & Yes \\
        ResNet50~\cite{he2016deep} & Random & Standard & No \\
        ResNet50~\cite{he2016deep} & ImageNet weights & Standard & Yes \\
        ResNet50~\cite{he2016deep} & ImageNet weights & Standard + Weight decay $1e-3$ & Yes \\
        DenseNet~\cite{huang2017densely} & Random & Standard & Yes \\
        EfficientNet-S~\cite{tan2019efficientnet} & Random & Standard & Yes \\
        EfficientNet-S~\cite{tan2019efficientnet} & ImageNet weights & Standard & Yes \\
    \bottomrule
    \end{tabular}
    \caption{Details of training configurations. Standard training procedure is described above. ImageNet weights are obtained from \texttt{torchvision}~\cite{torchvision2016} package.}
    \label{tab:training_configs}
\end{table}

\paragraph{Implementation details for baselines}

For COT, we used the suggestion of the authors to counter the cubic runtime of the method: we used batches of 2,500 images to estimate accuracy and averaged the accuracy over the batches, using up to 25,000 randomly sampled samples. To compute the Wasserstein distance we used the POT package~\cite{flamary2021pot} as suggested by the authors of COT~\cite{lu2023predicting} 

\newpage
\section*{Supp Note 3: Scatter plots Predicted versus True accuracy}
\begin{figure}[h!]
    \centering
    \includegraphics[width=\linewidth]{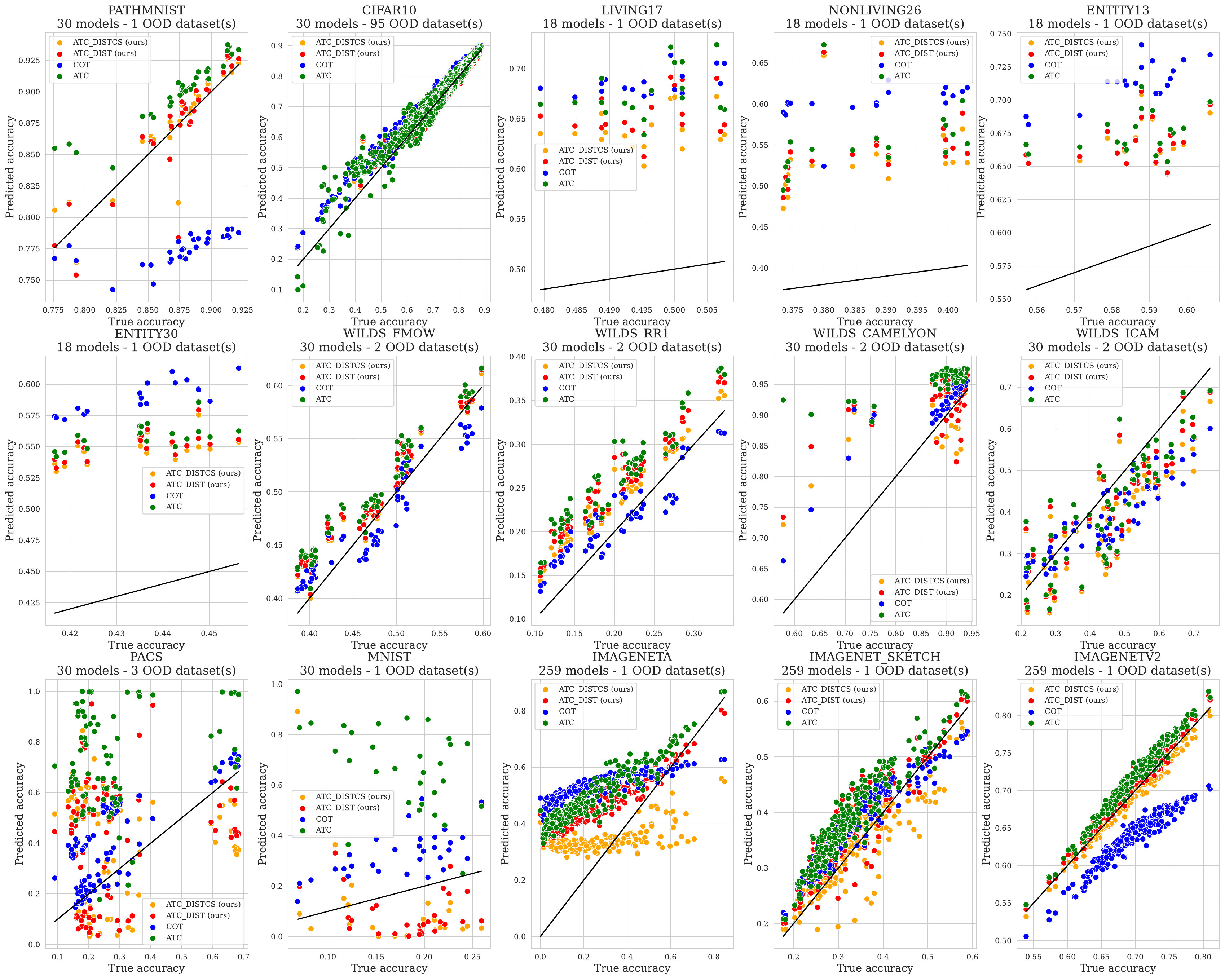}
    \caption{\textbf{Predicted versus true accuracy for all models and datasets.}}
    \label{fig:scatter}
\end{figure}

\newpage
\section*{Supp Note 4: Additional results for ablation study on the choice of distance estimation method.}
\begin{figure*}[h!]
    \centering
    \includegraphics[width=\linewidth]{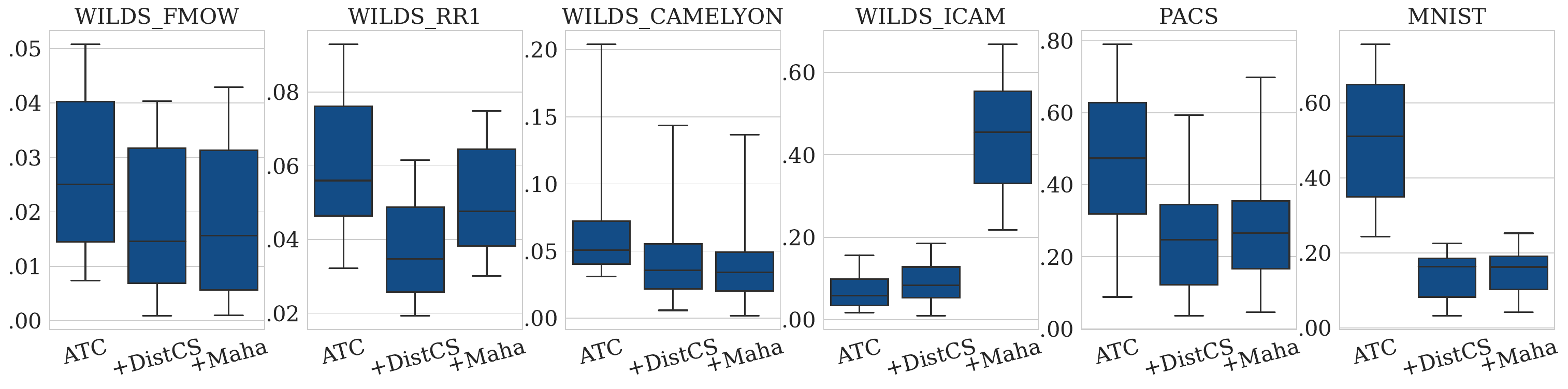}
    \caption{\textbf{Ablation study for the choice of distance estimation method: K-NN (DistCS) versus Mahalanobis distance (Maha).} Each boxplot shows the distribution of the Mean Absolute Error for accuracy estimation. Whiskers denote the [5\%;95\%]-percentiles of the distribution, outliers omitted for readability. Using distance improves the results for all but one dataset, no matter if K-NN or Mahalanobis distance. However, K-NN distance is better than Mahalanobis overall.}
    \label{fig:ablation_distance_2}
\end{figure*}
\section*{Supp Note 5: K-NN ablation study}
\begin{figure}[h!]
    \centering
    \includegraphics[width=\linewidth]{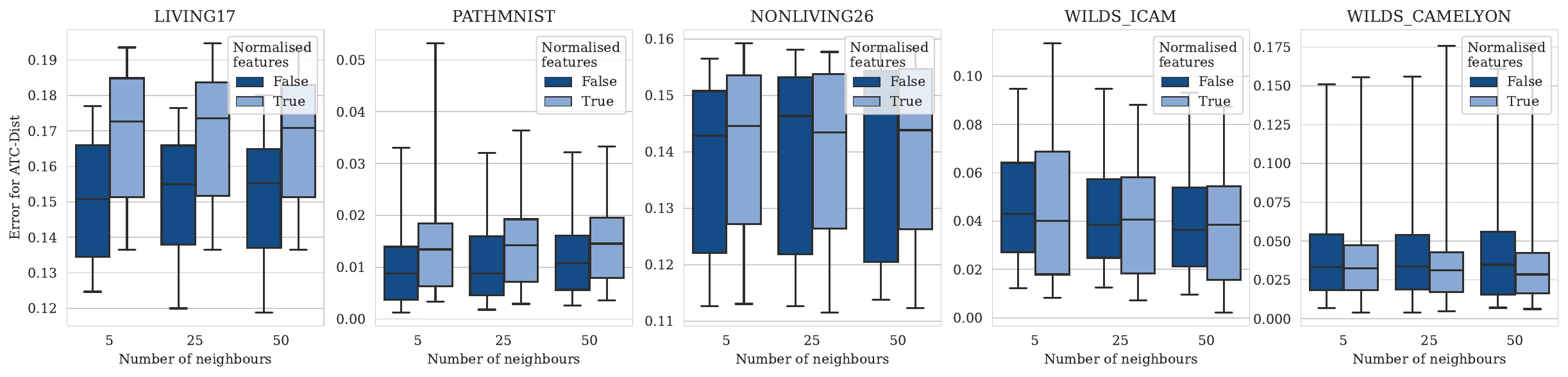}
    \caption{\textbf{Ablation study for parameters of nearest neighbours for ATCDist.} The method is not sensitive to the choice of number of neighbours and to normalisation of the features (i.e. dividing the features by their norm) does not significantly impact the performance. We compare the performance of the ATC-Dist estimator for 5 tasks in different settings. Each boxplot represents the distribution of absolute errors for accuracy estimation over all trained models. Whiskers denote the [5;95]$^{th}$ percentiles of the distribution. Outliers are omitted for readability.}
    \label{fig:ablation}
\end{figure}
\section*{Supp Note 6: Ablation study on the distance threshold choice}
\begin{figure}[h!]
    \centering
    \includegraphics[width=\linewidth]{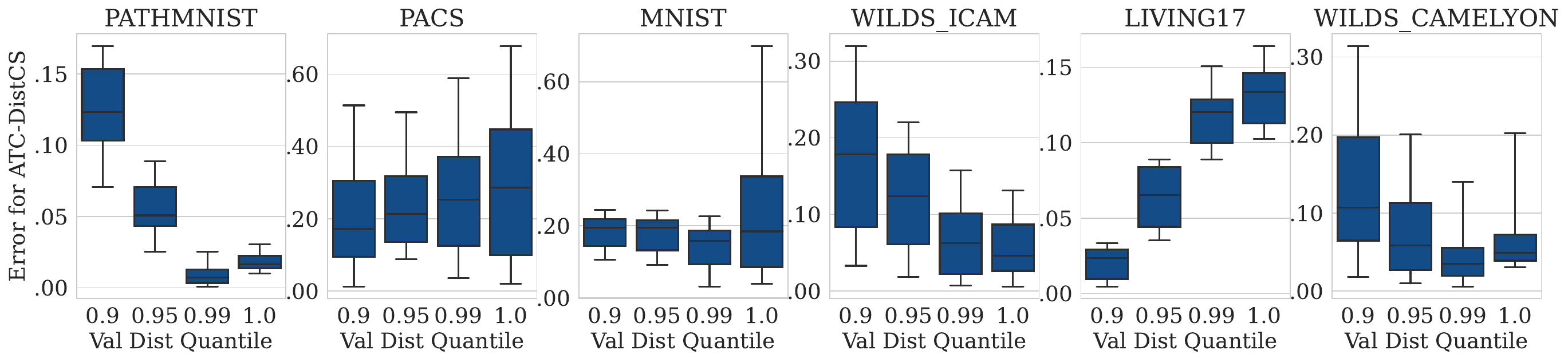}
    \caption{\textbf{Ablation study: MSE of ATC-Dist in function of distance threshold} (i.e. observed quantile on validation set). Our choice of threshold offers good generalisation across all tasks.}
    \label{fig:distance}
\end{figure}
\newpage

\end{document}